\documentclass[11pt,a4paper]{article}
\usepackage[hyperref]{eacl2021}
\usepackage{times}
\usepackage{latexsym}

\usepackage{amsmath}
\usepackage{graphicx}
\usepackage{booktabs}
\usepackage{multirow}
\usepackage{enumitem}
\usepackage{colortbl}
\usepackage{xcolor}
\usepackage{url}
\usepackage{amssymb}
\usepackage{amsthm}
\usepackage{mathtools}
\usepackage[ruled,vlined]{algorithm2e}
\usepackage{subcaption}
\usepackage{makecell}
\usepackage{xspace}
\usepackage{bbm}
\usepackage[ruled,vlined]{algorithm2e}
\usepackage[warn]{textcomp}
\usepackage{ marvosym }

\newcommand\blfootnote[1]{%
  \begingroup
  \renewcommand\thefootnote{}\footnote{#1}%
  \addtocounter{footnote}{-1}%
  \endgroup
}

\newcommand{\com}[1]{}

\usepackage{microtype}

\aclfinalcopy 
\def\aclpaperid{965} 

\title{Streaming Models for Joint Speech Recognition and Translation}

\author{Orion Weller$^{1}$*, Matthias Sperber$^{2}$,  Christian Gollan$^{2}$,  Joris Kluivers$^{2}$  \\ 
$^1$Brigham Young University \\
$^2$Apple\\
\normalsize{\texttt{oweller@byu.edu,\{m\char`_sperber,cgollan,jkluivers\}@apple.com}}
}

\begin{document}
\maketitle
\begin{abstract}
Using end-to-end models for speech translation (ST) has increasingly been the focus of the ST community.
These models condense the previously cascaded systems by directly converting sound waves into translated text. 
However, cascaded models have the advantage of including automatic speech recognition output, useful for a variety of practical ST systems that often display transcripts to the user alongside the translations. 
To bridge this gap, recent work has shown initial progress into the feasibility for end-to-end models to produce both of these outputs.
However, all previous work has only looked at this problem from the consecutive perspective, leaving uncertainty on whether these approaches are effective in the more challenging streaming setting.
We develop an end-to-end streaming ST model based on a re-translation approach and compare against standard cascading approaches.
We also introduce a novel inference method for the joint case, interleaving both transcript and translation in generation and removing the need to use separate decoders.
Our evaluation across a range of metrics capturing accuracy, latency, and consistency shows that our end-to-end models are statistically similar to cascading models, while having half the number of parameters.  We also find that both systems provide strong translation quality at low latency, keeping 99\% of consecutive quality at a lag of just under a second. \blfootnote{\textasteriskcentered Work done during an internship with Apple}
\end{abstract}

\section{Introduction}
Speech translation (ST) is the process of translating acoustic sound waves into text in a different language than was originally spoken in. 

This paper focuses on ST in a particular setting, as described by two characteristics: (1) We desire models that translate in a streaming fashion, where users desire the translation before the speaker has finished. This setting poses additional difficulties compared to consecutive translation, forcing systems to translate without knowing what the speaker will say in the future. (2) Furthermore, the speaker may want to verify that their speech is being processed correctly, intuitively seeing a streaming transcript while they speak \cite{Fugen2008,Hsiao2006}. For this reason, we consider models that produce both transcripts and translation jointly.\footnote{This corresponds to the \textit{mandatory transcript} case in the proposed categorization by \citet{Sperber2020}.} 

Previous approaches to streaming ST have typically utilized a cascaded system that pipelines the output of an automatic speech recognition (ASR) system through a machine translation (MT) model for the final result. These systems have been the preeminent strategy, taking the top place in recent streaming ST competitions \cite{pham2019iwslt,jan2019iwslt,elbayad2020trac,ansari-etal-2020-findings}.  Despite the strong performance of these cascaded systems, there are also some problems: error propagation from ASR output to MT input \cite{ruiz2014assessing}; ASR/MT training data mismatch and loss of access to prosodic/paralinguistic speech information at the translation stage \cite{Sperber2020}; and potentially sub-optimal latencies in the streaming context. End-to-end (E2E) models for ST have been proposed to remedy these problems, leveraging the simplicity of a single model to sidestep these issues. E2E models are also appealing from computational and engineering standpoints, reducing model complexity and decreasing parameter count.

Although initial research has explored E2E models for joint speech recognition and translation, no previous works have examined them in the streaming case, a crucial step in using them for many real-world applications. To understand this area more fully, we develop an E2E model to compare with its cascading counterpart in this simultaneous joint task. We build off the models proposed by \citet{sperber2020consistent} in the consecutive case, extending them for use in the streaming setting. We also use the re-translation technique introduced by \citet{niehues2018low} to maintain simplicity while streaming. To reduce model size, we introduce a new method for E2E inference, producing both transcript and translation in an interleaved fashion with one decoder. 

As this task requires a multi-faceted evaluation along several axes, we provide a suite of evaluations to highlight the differences of these major design decisions. This suite includes assessing translation quality, transcription quality, lag of the streaming process, output flicker, and consistency between the transcription and translation. We find that our E2E model performs similarly to the cascaded model, indicating that E2E networks are a feasible and promising direction for streaming ST.

\section{Proposed Method}
\paragraph{Network Architecture}
In the ST survey provided by \citet{sperber2020consistent}, they introduce several E2E models that could be used for the joint setting. As our work focuses on providing a simple but effective approach to streaming ST, we focus on the \textit{CONCAT} model, which generates both the transcript and translation in a concatenated fashion. We compare this E2E model against the standard cascading approach, following the architecture and hyperparameter choices used in \citet{sperber2020consistent}. All audio input models use the same multi-layer bidirectional LSTM architecture, stacking and downsampling the audio by a factor of three before processing. We note that although bidirectional encoders are unusual with standard ASR architectures, re-translation makes them possible. The cascaded model's textual encoder follows the architecture described in \citet{Vaswani2017} but replaces self-attention blocks with LSTMs. Decoder networks are similar, but use unidirectional LSTMs. More implementation details can be found in Appendix~\ref{app:models}.

\begin{figure}
   \centering
    \includegraphics[width=0.48\textwidth]{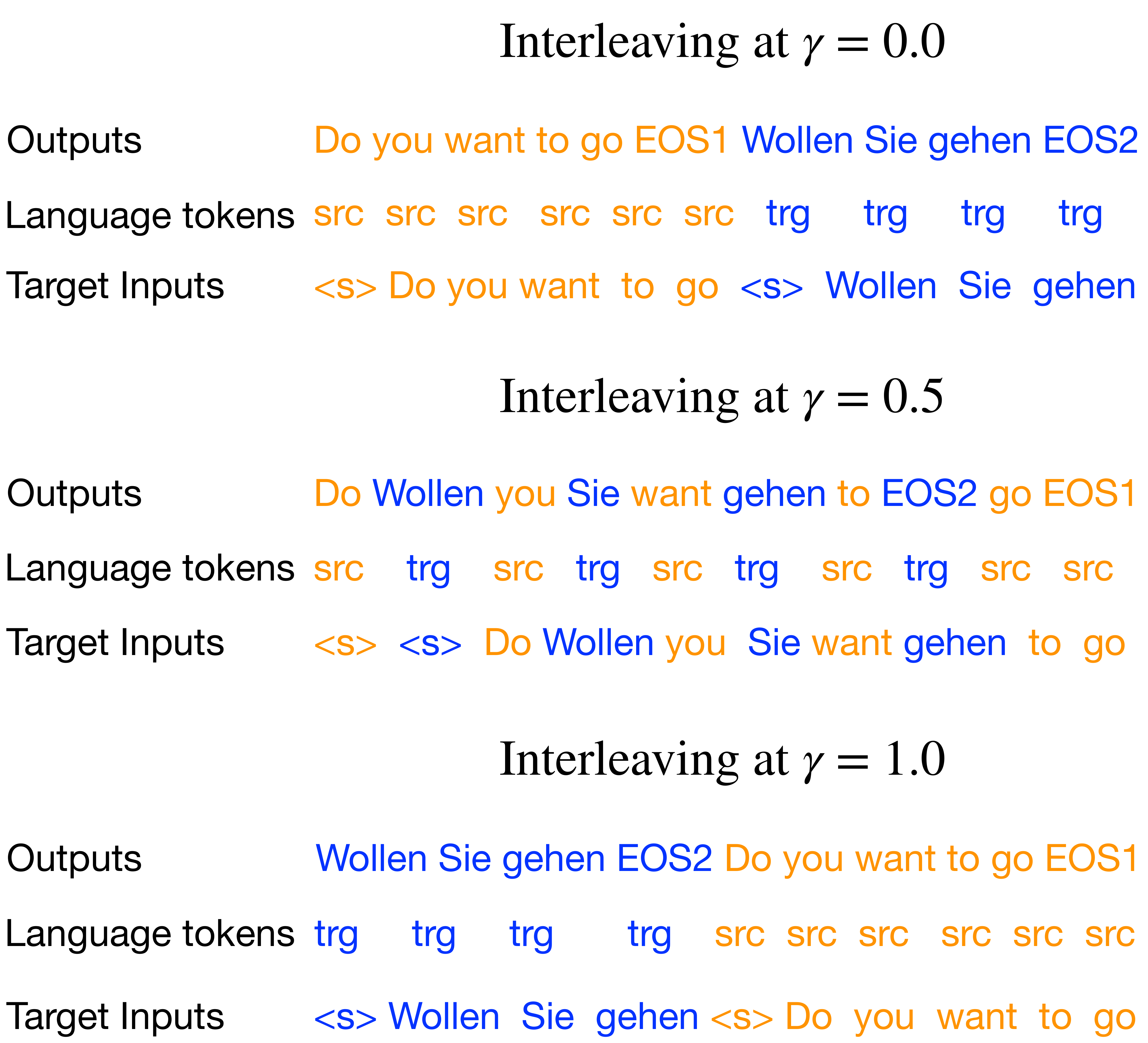}
    \caption{Example token representations (En\MVRightarrow De) for three different interleaving parameters (Section \ref{sec:interleave}). Language tokens indicate whether the data corresponds to the source transcript or the target translation and are used with a learned embedding that is summed with the word embeddings, as described in \citet{sperber2020consistent}.}
    \label{fig:interleaving}
\end{figure}

\label{sec:interleave}
In order to reduce model size and inference time for E2E networks, we introduce a novel method for interleaving both transcript and translation in generation, removing the need to use separate decoders. This method extends the \textit{CONCAT} model proposed by \citet{sperber2020consistent} to jointly decode according to the ratio given by the parameter $\gamma$ (Figure \ref{fig:interleaving}). When $\gamma=0.0$, we generate the transcript tokens until completion, followed by the translation tokens (vice versa for $\gamma=1.0$). At $\gamma=0.0$, our model is equivalent to the previously proposed model. Defining $\mathrm{count_i}$ as the count of $i$ tokens previously generated, transcription tokens as \textit{st} and translation tokens as \textit{tt}, we generate the next token as a transcription token if: \vspace{-1.3em}

\begin{align*}
    (1.0 - \gamma) * (1+\mathrm{count_{tt}}) > \gamma * (1+\mathrm{count_{st}}) 
\end{align*}

\noindent This approach enables us to produce tokens in an interleaving fashion, given the hyperparameter $\gamma$.

\begin{figure*}[t!]
    \centering
    \includegraphics[trim=0 10 10 5, clip,width=0.48\textwidth]{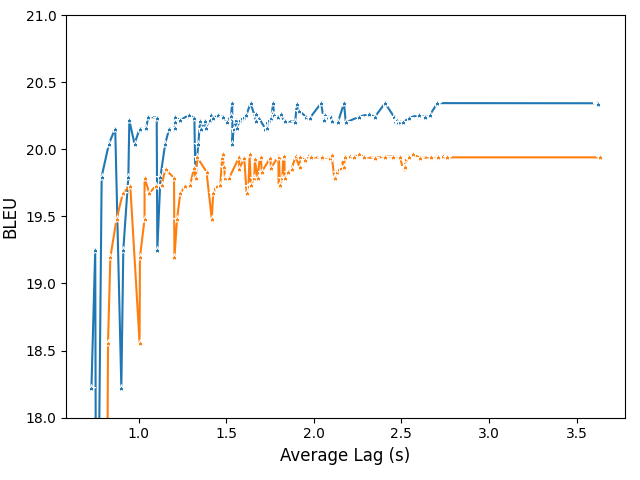}
    \includegraphics[trim=10 10 5 10, clip,width=0.48\textwidth]{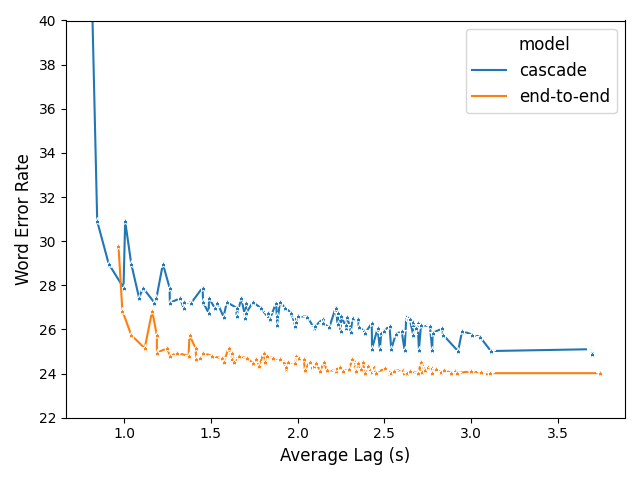}
    \caption{Left: average lag in seconds vs BLEU score. Right: average lag in seconds vs WER score. All points are the mean of each configuration's  score across the eight target languages. Configurations are the cross product of the values for $K$ and $F$, see Section~\ref{sec:inference}: Inference. Note that points near 1.0 AL have appx. 99\% of the unconstrained BLEU score. Results for the E2E model use $\gamma=0.5$.}.
    \label{fig:tradeoff}
\end{figure*}

\begin{table*}
\centering
\begin{tabular}{lllrrrrrrrrr}
\toprule
\textbf{Metric} & \textbf{Params} & \textbf{Model} & \textbf{De} & \textbf{Es} & \textbf{Fr} & \textbf{It} & \textbf{Nl} & \textbf{Pt} & \textbf{Ro} & \textbf{Ru} & \textbf{Average} \\
\midrule

BLEU $\uparrow$ & 217M & Cascade & \textbf{18.8} & \textbf{22.7} & \textbf{27.0} & \textbf{18.9} & \textbf{22.5} & \textbf{21.9} & \textbf{17.9} & \textbf{13.0} & \textbf{20.3} \\
& 107M & E2E  $\gamma$=0.0 & 18.1 & \textbf{23.1} & \textbf{27.0} & \textbf{18.7} & \textbf{22.3} & \textbf{22.2} & \textbf{17.6} & 12.2 & 20.2 \\
& 107M& E2E  $\gamma$=0.3 & 17.7 & 22.6 & 26.3 & 18.0 & 21.5 & 21.5 & 17.0 & 12.1 & 19.6 \\
& 107M& E2E  $\gamma$=0.5 & 18.2 & \textbf{22.8} & \textbf{27.0} & \textbf{18.6} &  21.9 & 21.9 & 17.1 & 12.0 & 19.9 \\
& 107M& E2E  $\gamma$=1.0 & 18.2 & \textbf{22.8} & \textbf{27.1} & \textbf{18.9} & \textbf{22.2} & \textbf{22.3} & \textbf{17.6} & \textbf{12.7} & \textbf{20.2}  \\
\midrule
WER $\downarrow$ & 217M & Cascade & 25.9 & 24.0 & 23.1 & 25.6 & 28.5 & 26.4 & 24.4 & \textbf{23.1} & 25.1  \\
& 107M & E2E  $\gamma$=0.0 & 24.2 & 23.5 & 23.3 & \textbf{23.0} & 23.4 & 25.3 & 24.1 & 23.6 & 23.8 \\
& 107M & E2E  $\gamma$=0.3 & 24.1 & 23.6 & 22.9 & 23.8 & 23.4 & 25.7 & 24.1 & 24.1 & 24.0 \\
& 107M & E2E  $\gamma$=0.5 & 24.5 & 23.9 & 22.9 & 23.8 & 23.4 & 25.7 & 24.3 & 23.6 & 24.0 \\
& 107M & E2E  $\gamma$=1.0 & \textbf{23.6} & \textbf{22.9} & \textbf{22.3} & \textbf{23.0} & \textbf{22.4} & \textbf{24.7} & \textbf{23.4} & \textbf{22.7} & \textbf{23.1} \\
\bottomrule
\end{tabular}
\caption{BLEU and WER scores for models trained on different target languages. Bold scores indicate results that are statistically similar to the best score using a bootstrap permutation test with $\alpha=0.05$.}
\label{tab:bleuwer}
\end{table*}

\paragraph{Re-translation} We use the re-translation method \cite{niehues2018low,arivazhagan2020re,arivazhagan2020respeech} as it provides a simple way to handle the streaming case. This method works by simply re-translating the utterance as new data arrives, updating its former prediction. As we are generating both transcript and translation, this avoids the challenging issue of combining the requirements for both components: streaming speech models need to manage the audio signal variability across time while streaming translation models need to overcome issues with reordering and lack of future context.

Alternative strategies to the re-translation approach include the chunk-based strategy explored by \citet{liu2020low}, which commits to all previous output chunks and \citet{ren2020simulspeech} who utilize an additional segmenter model trained via CTC \cite{graves2006connectionist} to create segments that are translated via wait-k \cite{ma2019stacl}. Although these approaches show effective results, they add additional complexity without addressing issues particular to streaming transcription.

\paragraph{Inference}
\label{sec:inference}
In order to generate quality-latency curves, we use several techniques to reduce latency and flicker at the cost of quality. The first is the mask-\textit{k} method proposed by \citet{arivazhagan2020respeech}, masking the last $K$ output tokens.  The second method is a form of constrained decoding: we define a hyperparameter $F$ that sets the number of free tokens allowed to change in the next re-translation. Thus, we constrain future output to match the first $\textit{len}(\textit{tokens}) - F$ tokens of the current output. All models use values $\{0, 1, 2, 3, 4, 5, 7, 10, 100\}$ for $K$ and $\{0, 1, 2, 3, 4, 5, 7, 10, 15, 20, 25, 100\}$ for $F$. For interleaving models, we set $K$ and $F$ on both transcript and translation tokens.

\begin{table*}
\centering
\hspace*{-0.25cm}
\begin{tabular}{lrrrrrr}
\toprule
\textbf{Model} & \textbf{En\MVRightarrow De Incr.} & \textbf{En\MVRightarrow De Full} & \textbf{En\MVRightarrow Es Incr.} & \textbf{En\MVRightarrow Es Full} & \textbf{Mean Incr.} & \textbf{Mean Full} \\
\midrule
Cascade & \textbf{13.8} & \textbf{13.2} & \textbf{12.2} & \textbf{11.6} & \textbf{14.1} & \textbf{13.4} \\
Concat  $\gamma$=0.0 & 17.6 & 16.7 & 14.9 & 13.8 & 17.0 & 16.0 \\
Concat  $\gamma$=0.3 & 17.2 & 16.6 & 14.3 & 13.7 & 16.6 & 15.8  \\
Concat  $\gamma$=0.5 & 17.8 & 16.5 & 14.8 & 13.3 & 17.3 & 15.7  \\
Concat  $\gamma$=1.0 & 17.3 & 16.8 & 14.9 & 13.7 & 16.9 & 15.8 \\

\bottomrule
\end{tabular}
\caption{Consistency scores for En\MVRightarrow De, En\MVRightarrow Es, and average results over all languages; lower is better (see \citet{sperber2020consistent}). \emph{Incr.} stands for the incremental consistency score, or the average consistency throughout re-translation. Bold scores indicate results that are statistically similar to the best score using a bootstrap permutation test with $\alpha=0.05$.}
\label{tab:consistency}
\end{table*}

\section{Experimental Settings}
\paragraph{Data} We use the MuST-C corpus \cite{DiGangi2019c} since it is the largest publicly available ST corpus, consisting of TED talks with their English transcripts and translations into eight other language pairs. The dataset consists of at least 385 hours of audio for each target language. 

We utilize the log Mel filterbank speech features provided with the corpus as input for the ASR and E2E models. To prepare the textual data, we remove non-speech artifacts (e.g. ``(laughter)" and speaker identification) and perform subword tokenization using SentencePiece \cite{Kudo2018a} on the unigram setting. Following previous work for E2E ST models, we use a relatively small vocabulary and share transcription and translation vocabularies. We use MuST-C dev for validation and report results on tst-COMMON, utilizing the segments provided (Appendix \ref{app:segments}).

\label{sec:training}
\paragraph{Prefix Sampling} We implement techniques developed by \citet{niehues2018low,arivazhagan2020respeech} for improving streaming ST, sampling a random proportion of each training instance as additional data to teach our models to work with partial input.  See Appendix~\ref{app:prefix} for implementation details.

\paragraph{Metrics} We evaluate these models on a comprehensive suite of metrics: sacrebleu (\emph{BLEU}, \citet{Post2019}) for translation quality, word error rate (\emph{WER}, \cite{fiscus1997post}) for transcription quality, average lag (\emph{AL}, \citet{ma2019stacl}) for the lag between model input and output, and normalized erasure (\emph{NE}, \citet{arivazhagan2020re}) for output flicker. Measuring consistency is a nascent area of research; we use the robust and simple lexical consistency metric defined by \citet{sperber2020consistent}, which uses word-level translation probabilities.  To show how consistent these results are while streaming, we compute an incremental consistency score, averaging the consistency of each re-translation.

\section{Results}
Results for the quality-latency curves created by the use of constrained decoding and mask-\textit{k} (Section~\ref{sec:training}) are shown in Figure~\ref{fig:tradeoff}. Unconstrained settings are used for all results in table form. For convenience, bold scores indicate the highest performing models in each metric according to a bootstrap permutation test.

\paragraph{Translation Quality} We see in Table~\ref{tab:bleuwer} that the cascaded model slightly outperforms some E2E models, while achieving statistically similar performance to the $\gamma=1.0$ model. We note however, that the cascaded model has nearly twice as many parameters as the E2E models (217M vs 107M). When we examine these models under a variety of different inference conditions (using constrained decoding and mask-\textit{k} as in \citet{arivazhagan2020re}), we further see this trend illustrated through the quality vs latency trade-off (left of Figure~\ref{fig:tradeoff}), with both models retaining 99\% of their BLEU at less than 1.0 AL.

\paragraph{Transcription Quality} Conversely, Table~\ref{tab:bleuwer} and the right of Figure~\ref{fig:tradeoff} show that the $\gamma=1.0$ E2E model performs similarly or slightly better than the cascaded model across all inference parameters and all target languages. With an AL of 1.5, the E2E model loses only 3\% of its performance.

\paragraph{Consistency} The E2E models perform worse than the cascaded on consistency, with the best models being approximately 18\% less consistent (Table~\ref{tab:consistency}). The cascaded model also maintains better scores through each re-translation (\emph{Incr.}).\footnote{Initial experiments indicate that the triangle E2E architecture  \cite{sperber2020consistent} model may perform better on consistency in our streaming setting, but due to time constraints we were not able to explore this further. Future work exploring alternative architectures or decoding techniques \cite{le-etal-2020-dual} may provide fruitful avenues of research.}

\paragraph{Flicker} We note that the flicker scores for cascade and E2E models are similar, with both having normalized erasure scores of less than 1 and the majority of inference settings having less than the ``few-revision" threshold of 0.2 (proposed by \citet{arivazhagan2020re}). More NE details are found in Appendix~\ref{app:ne}.

\paragraph{Interleaving Rate} Table~\ref{tab:bleuwer} also shows us the overall results for different interleaving rates. We see that interleaving at a rate of 1.0 has the best quality scores (0.7 less WER than the next best rate, the base $\gamma=0.0$ model) but the worst consistency (Table~\ref{tab:consistency}). Conversely, $\gamma=0.3$ has the worst quality scores but the best consistency.

\section{Conclusion}
We focus on the task of streaming speech translation, producing both a target translation and a source transcript from an audio source. We develop an end-to-end model to avoid problems that arise from the use of cascaded models for streaming ST.
We further introduce a new method for joint inference for end-to-end models, generating both translation and transcription tokens concurrently.  We show that our novel end-to-end model, with only half the number of parameters, is comparable to standard cascaded models across a variety of evaluation categories: transcript and translation quality, lag of streaming, consistency between transcript and translation, and re-translation flicker.
We hope that this will spur increased interest in using end-to-end models for practical applications of streaming speech translation.

\bibliography{eacl2021,library}
\bibliographystyle{acl_natbib}

\appendix

\section{Model Details}
\label{app:models}
\begin{figure}[b!]
   \centering
    \includegraphics[width=0.48\textwidth]{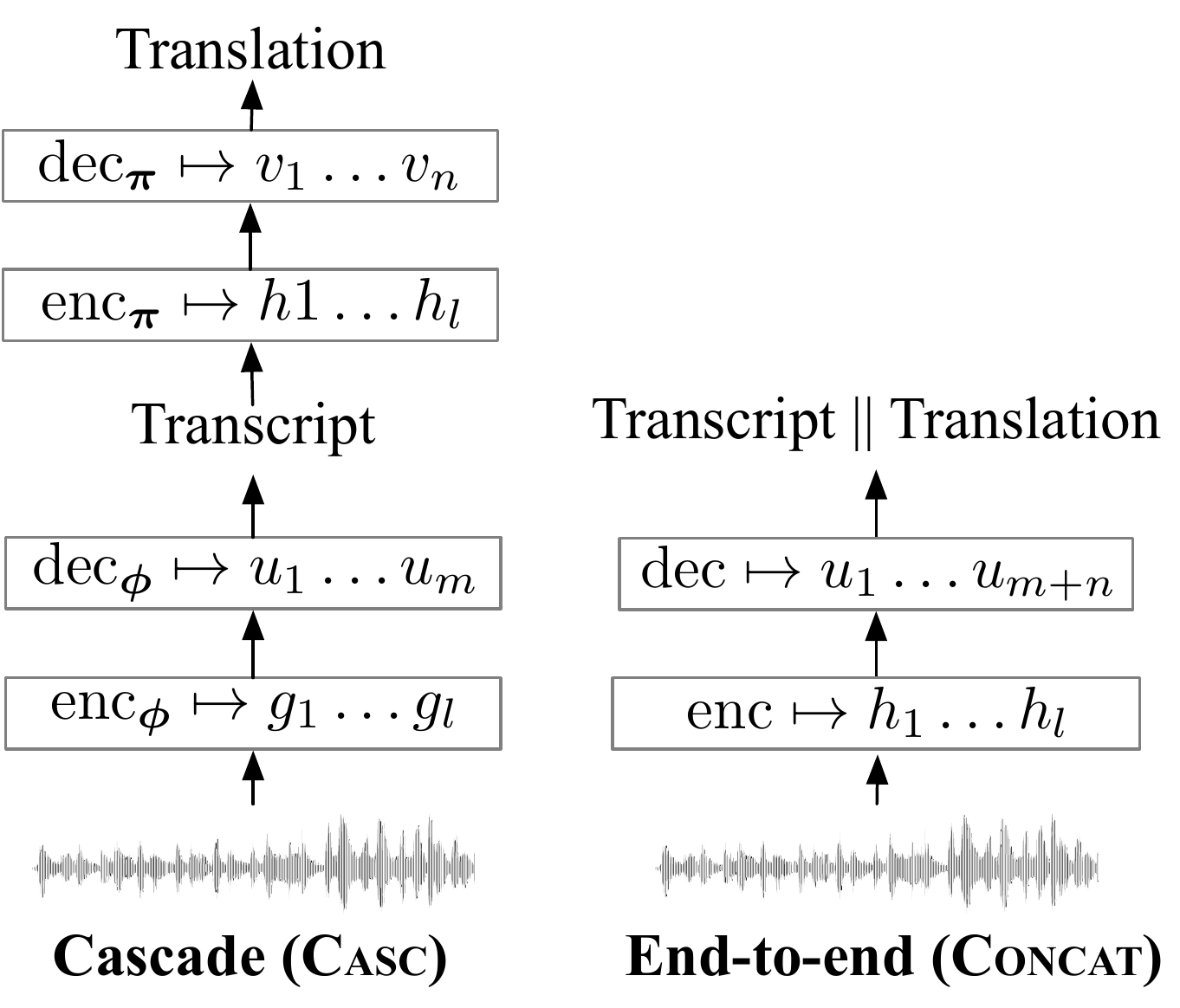}
    \caption{Architectures of the cascade and concatenated model}
    \label{fig:arch}
\end{figure}

In this section we will describe implementation details of the model architectures (shown in Figure \ref{fig:arch}) and training processes.

\paragraph{Model Architectures}
Unless otherwise noted, the same hyperparameters are used for all models. Weights for the speech encoder are initialized based on a pre-trained attentional ASR task that is identical to the ASR part of the direct multitask model. Other weights are initialized according to \citet{Glorot2010}. The speech encoder is a 5-layer bidirectional LSTM with 700 dimensions per direction. Attentional decoders consist of 2 Transformer blocks \cite{Vaswani2017} but use 1024-dimensional unidirectional LSTMs instead of self-attention, except for the \textit{CONCAT} model, which uses 3 layers. 

For the cascade’s MT model, encoder/decoder both contain 6 layers with 1024-dimensional LSTMs. Subword embeddings are size 1024. We regularize using LSTM dropout with $p=0.3$, decoder input word-type dropout \cite{Gal2015}, and attention dropout, both $p=0.1$. We apply label smoothing with strength $\epsilon=0.1$.

\paragraph{Training}
We optimize using Adam \cite{Kingma2014} with $\alpha=0.0005$, $\beta_{1} =0.9$, $\beta_{2}=0.98$, 4000 warm-up steps, and learning rate decay by using the inverse square root of the iteration of each instance. We set the batch size dynamically based on the sentence length, such that the average batch size is 128 utterances. The training is stopped when the validation score has not improved over 10 epochs, where the validation score is corpus-level translation BLEU score (for the E2E and MT models) and corpus-level WER for the cascade's ASR model.

\begin{figure}[t!]
    \hbox{
        \hspace{-1em} \includegraphics[width=0.5\textwidth]{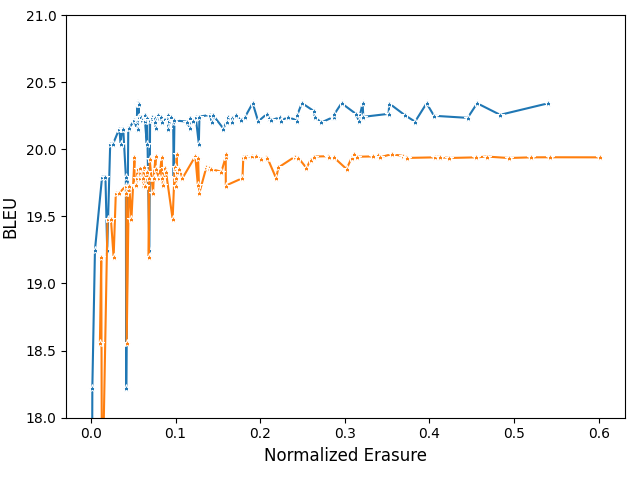}
    }
    \hbox{
        \hspace{-1em} \includegraphics[width=0.5\textwidth]{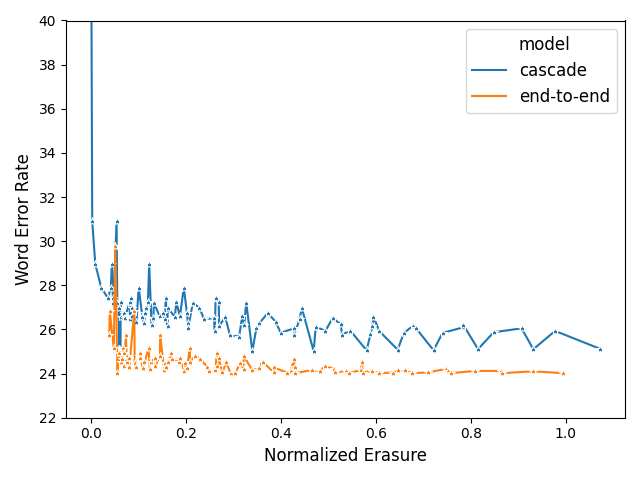}
    }
    \caption{Left: average lag in seconds vs NE score. Right: NE vs WER score. All values are the mean of the results from the eight target languages.}
    \label{fig:ne}
\end{figure}

For decoding and generating n-best lists, we use beam size 5 and polynomial length normalization with exponent 1.5. Our implementation is based on PyTorch \cite{Paszke2019} and XNMT \cite{Neubig18xnmt}, and all models are trained in single-GPU environments, employing Tesla V100 GPUs with 32 GB memory.  Most E2E and ASR models converged after approximately 30 epochs or 5 days of training. MT models converged after approximately 50 epochs or 2 days of training.

\section{Normalized Erasure (Output Flicker)}
\label{app:ne}
We see similar curves for both the cascaded model and the E2E model when comparing normalized erasure in Figure~\ref{fig:ne}. We see that most settings have an NE score of less than 0.2, while virtually all settings are less than 1.  We note that a proportion of 0.2 for NE means that, on average, 1/5 of the tokens change once before they settle to their final state.

\section{Prefix Training}
\label{app:prefix}
We used prefix training to increase stability and reduce flickering in the streaming setting.  We conducted this by utilizing each training instance twice in each epoch: one as normal and the other with only the prefix.  The length of the prefixes were randomly sampled from [0, 1].  We found that this additional data augmentation was particularly helpful; without it, the models would hallucinate the rest of a partial sentence.

We further found that starting the prefix sampling data augmentation too late in training was also negative. After testing initial models on the dev set, we found that starting this additional augmentation 15 epochs after training was best.

\section{Utterance Segmentation}
\label{app:segments}
We follow the audio segments provided in the MuST-C corpus, created through a use of human alignment and XNMT \cite{Neubig18xnmt}. We note that there exist a variety of methods for creating segments for such models, however, we leave additional exploration of E2E alignment methods as future work.

\end{document}